\documentclass[conference]{IEEEtran}
\IEEEoverridecommandlockouts

\usepackage{cite}

\usepackage{amsmath,amssymb,amsfonts}
\usepackage{algorithmic}
\usepackage{graphicx}
\usepackage{textcomp}
\usepackage{xcolor}
\usepackage{float}
\usepackage{booktabs}
\usepackage{lmodern}
\usepackage{tikz}
\usetikzlibrary{shapes,arrows,backgrounds,fit}
\def\BibTeX{{\rm B\kern-.05em{\sc i\kern-.025em b}\kern-.08em
    T\kern-.1667em\lower.7ex\hbox{E}\kern-.125emX}}
\begin{document}

\title{LiZIP: An Auto-Regressive Compression Framework for LiDAR Point Clouds\\
\thanks{This work was supported by the Dubai RDI Grant Program under the project 'CogWaters' (Project ID: 2025/DRDI0163).}
}

\author{\IEEEauthorblockN{Aditya Shibu}
\IEEEauthorblockA{\textit{MACS} \\
\textit{Heriot Watt University}\\
Dubai, United Arab Emirates \\
as2397@hw.ac.uk}
\and
\IEEEauthorblockN{Kayvan Karim}
\IEEEauthorblockA{\textit{MACS} \\
\textit{Heriot Watt University}\\
Dubai, United Arab Emirates \\
k.karim@hw.ac.uk}
\and
\IEEEauthorblockN{Claudio Zito}
\IEEEauthorblockA{\textit{MACS} \\
\textit{Heriot Watt University}\\
Dubai, United Arab Emirates \\
c.zito@hw.ac.uk}
% \and
% \and
% \and
}

\maketitle

% \begin{abstract}
% This paper proposes LiZIP, a lightweight deep learning-based framework for the lossless compression of LiDAR point clouds. The methodology employs a neural predictive coding model, where a Multi-Layer Perceptron (MLP) is trained to predict a point's coordinates based on its neighbouring local context. The system achieves compression by encoding only the minimal residuals between the predicted and actual coordinates. This paper presents a quantitative evaluation of LiZIP on benchmark datasets and against established baselines. specifically Google Draco, GZip and LasZip, demonstrating its superior compression ratios and computational efficiency compared to traditional methods. The results indicate that LiZIP effectively reduces storage requirements while maintaining perfect integrity (0.0 reconstruction error) of the original point cloud data, making it a promising solution for applications in autonomous driving, robotics, and 3D mapping. The final results aim to provide a definitive answer on the feasibility of this deep learning approach for real-time applications in next-generation autonomous systems.
% \end{abstract}

\begin{abstract}
The massive volume of data generated by LiDAR sensors in autonomous vehicles creates a bottleneck for real-time processing and vehicle-to-everything (V2X) transmission. Existing lossless compression methods often force a trade-off: industry-standard algorithms (e.g., LASzip) lack adaptability, while deep learning approaches suffer from prohibitive computational costs. This paper proposes LiZIP, a lightweight, near-lossless zero-drift compression framework based on neural predictive coding. By utilizing a compact Multi-Layer Perceptron (MLP) to predict point coordinates from local context, LiZIP efficiently encodes only the sparse residuals.

We evaluate LiZIP on the NuScenes and Argoverse datasets, benchmarking against GZip, LASzip, and Google Draco (configured with 24-bit quantization to serve as a high-precision geometric baseline). Results demonstrate that LiZIP consistently achieves superior compression ratios across varying environments. The proposed system achieves a 7.5\%--14.8\% reduction in file size compared to the industry-standard LASzip and outperforms Google Draco by 8.8\%--11.3\% across diverse datasets. Furthermore, the system demonstrates generalization capabilities on the unseen Argoverse dataset without retraining. Against the general purpose GZip algorithm, LiZIP achieves a reduction of 38\%--48\%. This efficiency offers a distinct advantage for bandwidth-constrained V2X applications and large-scale cloud archival.
\end{abstract}


\section{Introduction}
Three-dimensional (3D) point clouds have become an essential format for representing geometric information in modern intelligent systems \cite{Liu2021PointCloudAnalysis}. Acquired primarily through Light Detection and Ranging (LiDAR) sensors, these point clouds provide precise, reliable depth information essential for applications ranging from robotics to autonomous driving \cite{Liu2021PointCloudAnalysis, Biasizzo2013FPGACompression, Zhou2017VoxelNet, Li2020LidarAutonomousDriving}. Unlike 2D images, which project the world onto a plane, LiDAR data offers 3D spatial information that remains reliable even in adverse lighting conditions \cite{Liu2021PointCloudAnalysis}. However, this comes at a significant cost that high-resolution LiDAR sensors generate hundreds of thousands of points per second, creating massive data volumes that can strain storage systems and transmissions \cite{Cao2025RealtimeLiDARCompression}.

This bottleneck is critical for Connected and Automated Vehicles (CAVs) evolving toward Level 4 and Level 5 automation \cite{Shladover2018ConnectedVehicles, Yang2023LiDARLLM}. The unstructured nature of point clouds makes them difficult to compress using standard methods like JPEG or H.265 \cite{Liu2021PointCloudAnalysis, Gupta2017LosslessCompression}, necessitating specialized compression for real-time edge computing \cite{Cao2025RealtimeLiDARCompression}.

Current industry standards for lossless compression such as LASZip, rely on predictive coding schemes that estimate point attributes based on previous points and encode these residuals using arithmetic coding. While LASzip is highly efficient for specific flight-line data structures, it relies on handcrafted prediction rules and fixed mathematical models that may not fully exploit the complex local geometric correlations present in dense urban LiDAR scans \cite{Isenburg2013LASzip}. Other approaches, such as Google Draco, utilize KD-tree spatial subdivision, which can be computationally intensive and is often optimized for lossy quantization rather than lossless fidelity \cite{Google2017Draco, Mongus2012LiDARCompressionComparison}.

% To address these limitations, we propose LiZIP, a novel deep learning framework for near-lossless LiDAR compression based on Neural Predictive Coding. Departing from traditional handcrafted predictors, LiZIP utilizes a lightweight Multi-Layer Perceptron (MLP) to learn the local geometric structure of the point cloud. By implementing a lightweight voxel-based spatial sorting, we organize the raw, unordered data into structured local contexts. This allows the MLP to accurately predict a target point's coordinates based on its spatially sorted neighbours. Rather than storing raw floating-point coordinates, we compress the residuals, the difference between the actual and predicted positions, which are significantly smaller and possess lower entropy.

We implemented our framework in PyTorch, utilizing Mean Squared Error (MSE) as the loss function to minimize prediction errors during the training of the neural network \cite{Elharrouss2025LossFunctions, Nwankpa2018ActivationFunctions}. By focusing on the residual encoding, LiZIP aims to bridge the gap between heavy, high-latency deep-learning models and traditional, rule-based compressors. In this paper, we evaluate LiZIP against industry benchmarks including LASzip, Google Draco, and GZip, demonstrating its potential as a scalable solution for the data-intensive demands of next-generation autonomous systems.

The remainder of this paper is organized as follows: Section II reviews related work in point cloud compression. Section III details the architecture of the LiZIP compressor and the neural predictive model. Section IV presents the experimental setup and ablation studies, followed by benchmarking results in Section V. Finally, Section VI offers discussions along with future works on how to improve the system and finally Section VII concludes the paper.

\section{Related Work}

Managing LiDAR point clouds requires compression algorithms that balance compression ratios with computational latency. Existing approaches: traditional geometric methods, general-purpose entropy coding, and deep learning frameworks, each have distinct advantages but struggle with this real-time trade-off.

\subsection{Traditional and Geometric Methods}

The current industry standard for lossless compression is LASzip, which employs a hand-crafted predictive coding scheme. LASzip operates by predicting the coordinates of a point based on previous points and encoding the residual errors using arithmetic coding \cite{Isenburg2013LASzip}. While highly effective for linearly correlated data (such as flight-line aerial scans), its reliance on fixed mathematical prediction rules limits its adaptability to the complex, irregular geometries found in terrestrial autonomous driving datasets. Our experiments demonstrate that LiZIP outperforms LASzip, reducing file sizes by 11.2\% on average, indicating that a learnable neural predictor captures local geometric correlations more effectively than static rules.

Geometric decomposition methods, such as Google Draco, represent another dominant class of compression algorithms. Draco utilizes KD-trees (k-dimensional trees) to spatially partition point clouds, encoding the topology and quantizing attribute values \cite{Google2017Draco}. Similar octree-based approaches, such as VoxelContext-Net \cite{Que2021VoxelContextNet} and OctSqueeze \cite{Huang2021OctSqueeze}, organize data into hierarchical structures to exploit spatial redundancy. However, the construction and traversal of these tree structures introduce significant computational overhead during encoding and decoding. In the context of Vehicle-to-Everything (V2X) communication, this latency is critical. Our results show that LiZIP is not only approximately 10\% smaller than Draco-compressed files, proving that neural predictive coding offers a more bandwidth-efficient solution than heavy geometric partitioning.

\subsection{General-Purpose Compression}

General-purpose lossless algorithms such as GZip (Deflate) and LZMA, treat LiDAR point clouds as one-dimensional byte streams. These methods utilize dictionary-based schemes (e.g., LZ77) and entropy encoding (e.g., Huffman coding) to remove statistical redundancy \cite{Gupta2017LosslessCompression, Kotb2018AirborneLiDARCompression, Ng1998CompressionTechniques}. However these algorithms are fundamentally agnostic to the underlying 3D spatial structure of the data. As point clouds are inherently sparse and unordered \cite{Qi2017PointNet}, adjacent bytes in the raw binary file may represent points that are distant in 3D Euclidean space. Consequently, general-purpose methods fail to exploit local geometric correlations. By implementing Morton Sorting (Z-order curve) for spatial sorting prior to neural prediction, LiZIP creates a highly correlated sequence that generic compressors cannot achieve, resulting in a 48\% improvement in compression ratio over GZip.

\subsection{Deep Learning-Based Approaches}

The introduction of deep learning on point sets, pioneered by PointNet \cite{Qi2017PointNet} and VoxelNet \cite{Zhou2017VoxelNet}, has introduced powerful capabilities for feature extraction and semantic understanding. PointNet processes raw point clouds directly to learn global features \cite{Qi2017PointNet}, while VoxelNet partitions space into voxels and applies 3D convolutions to extract volumetric features \cite{Zhou2017VoxelNet}. Due to the inherently sparse nature of LiDAR data, specialized sparse convolutional networks have been developed to efficiently process 3D point clouds while reducing computational overhead \cite{Luo2024SparseConvolutionGPU}. While these architectures excel at tasks like object detection and classification, they are not suited for lossless compression due to two primary reasons.

First, standard deep learning models are computationally intensive. Voxel-based methods suffer from the curse of dimensionality and high memory usage due to the sparsity of LiDAR data \cite{Zhou2017VoxelNet}, often requiring powerful GPUs for inference. In contrast, LiZIP utilizes a lightweight Multi-Layer Perceptron (MLP) with only approximately 540 KB of weights. This lean architecture allows LiZIP to perform inference on a CPU in approximately 173 ms, making it viable for resource-constrained computing in autonomous vehicles without hardware acceleration.

Second, most deep learning compression frameworks, such as MNet \cite{Nguyen2023PointCloudCompression} or autoencoder-based approaches, focus on lossy compression or attribute compression (e.g., color, intensity) rather than precise geometry. They often introduce quantization artifacts or downsampling to fit data within regular grids \cite{Que2021VoxelContextNet}. While recent neural methods like OctSqueeze report high density, they often require GPU acceleration and lack publicly available, optimized C++ implementations. RENO \cite{You2025RENO}, for instance, achieves strong compression through learned entropy models but depends on GPU execution for both encoding and decoding, limiting deployment in GPU-constrained automotive systems. The overhead of maintaining complex tree structures or heavyweight neural architectures often degrades runtime performance, particularly during decoding when rapid reconstruction is critical for downstream perception tasks. Our evaluation focuses on CPU-based industry standards (LASzip, Draco) to demonstrate LiZIP's practical viability for onboard automotive systems. LiZIP addresses this gap by strictly enforcing a near-lossless constraint: The MLP is used solely to generate a prediction, and the exact residuals (Actual - Predicted) are encoded. This hybrid approach leverages the adaptability of deep learning while guaranteeing the data fidelity required for safety-critical applications.

\section{The LiZIP compressor}
\begin{figure*}[t]
\centering
\begin{tikzpicture}[node distance=1.5cm, auto,
    block/.style={rectangle, draw, fill=blue!10, text width=5em, text centered, rounded corners, minimum height=3em},
    wideblock/.style={rectangle, draw, fill=blue!10, text width=7em, text centered, rounded corners, minimum height=3em},
    line/.style={draw, -latex', thick},
    cloud/.style={draw, ellipse, fill=red!10, node distance=2.5cm, minimum height=2em},
    container/.style={draw, dashed, inner sep=0.3cm, fill=green!5, label={above:\textbf{Parallel Block Processing}}}]

    \node [cloud] (input) {Raw LiDAR};
    \node [wideblock, right of=input, node distance=3.2cm] (sort) {Spatial Sorting (Morton)};
    \node [wideblock, right of=sort, node distance=3.2cm] (quant) {Quantization (Float $\rightarrow$ Int)};
    \node [wideblock, right of=quant, node distance=4.0cm] (context) {Context Generation ($k=3$)};
    \node [wideblock, right of=context, node distance=3.8cm] (mlp) {Neural Predictor (MLP)};

    \node [wideblock, above of=context, node distance=2.2cm] (anchors) {Extract Anchors ($k=3$)};

    \node [wideblock, below of=mlp, node distance=2.5cm] (diff) {Residual Calculation};

    \begin{pgfonlayer}{background}
        \node [container, fit=(anchors) (context) (mlp) (diff)] (par_block) {};
    \end{pgfonlayer}

    \node [wideblock, left of=diff, node distance=3.8cm] (shuffle) {Byte Shuffling};
    \node [wideblock, left of=shuffle, node distance=4.0cm] (lzma) {Entropy Coding (LZMA/ZLib)};
    \node [cloud, left of=lzma, node distance=3.7cm] (output) {Bitstream (.lizip)};

    \path [line] (input) -- (sort);
    \path [line] (sort) -- (quant);
    \path [line] (quant) -- (context);

    \path [line] (quant) |- (anchors);
    \path [line] (anchors.east) -- ++(0.8,0) -- ++(0,-4.3) -- ([yshift=0.4cm]shuffle.east);
    \path [line, dashed] (anchors) -- node [right] {Init} (context);

    \path [line] (context) -- (mlp);
    \path [line] (mlp) -- node [midway, right] {Pred} (diff);
    \path [line] (diff) -- (shuffle);
    \path [line] (shuffle) -- (lzma);
    \path [line] (lzma) -- (output);

    \path [line, dashed] (quant.south) -- ++(0,-1.0) -| node [pos=0.08, above] {Actual $P_t$} ([xshift=-0.5cm]diff.north);

    \path [line, dashed] (diff.south) -- ++(0,-0.5) -- ++(-2.0,0) -- ++(0,3.15) node [pos=0.8, left] {History} -- ++(-0.3,0) -- ([yshift=-0.4cm]context.east);

\end{tikzpicture}
\caption{System Architecture of the LiZIP Compressor. Raw points are sorted and quantized. Processing occurs in \textbf{parallel blocks}: the first $k$ points (Anchors) initialize the context, while the MLP predicts the rest. Both anchors and residuals are byte-shuffled and entropy coded.}
\label{fig:pipeline}
\end{figure*}
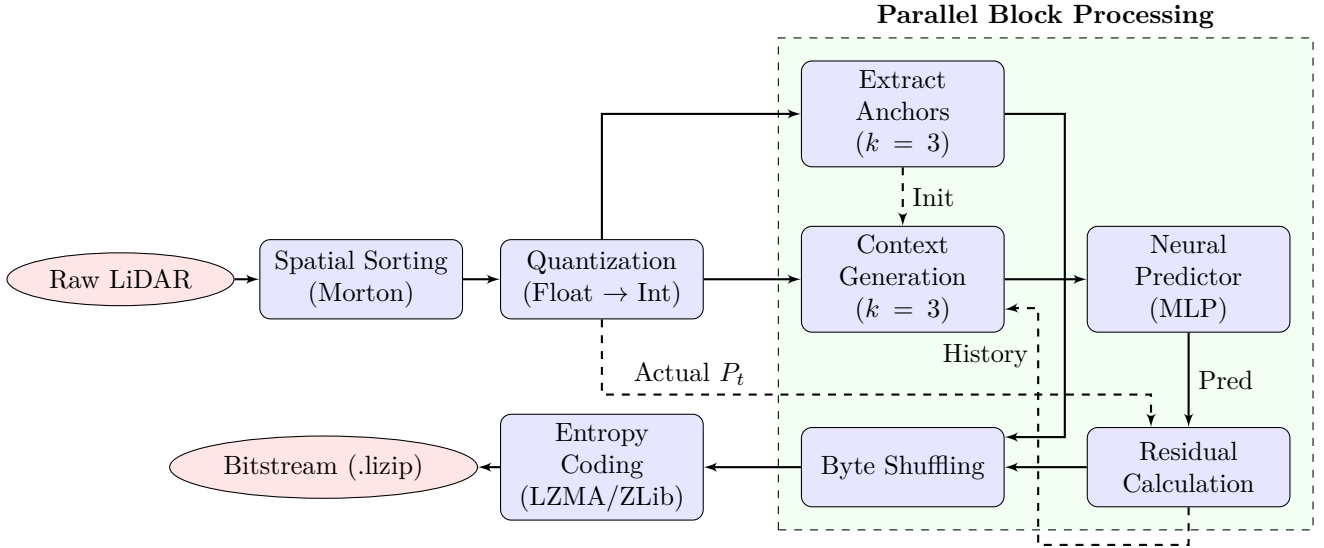

The LiZIP compressor does not rely on hand-crafted prediction rules like LASzip. Instead, it views the LiDAR point cloud as a continuous sequence of spatial coordinates that can be represented by a non-linear function. The compressor encodes points in chunks to maximize memory locality and enable parallel decoding in future iterations. When starting a new chunk, the LiZIP compressor initializes the spatial index and the neural weights. All following points are compressed point by point using the pipeline illustrated in Figure \ref{fig:pipeline}.

\subsection{Spatial Organization (Morton Sorting)}
LiDAR data often arrives in acquisition order (e.g., scan rings), which is suboptimal for predictive coding. To maximize spatial autocorrelation, LiZIP maps every 3D point $P = (x, y, z)$ to a single 64-bit integer using a Morton Code (Z-Curve). This code interleaves the bits of the coordinate axes, ensuring that points close in the 1D sequence are physically adjacent in 3D space \cite{Wei2020MortonCodePCC}. The compressor sorts the chunk based on these codes before prediction begins.

Formally, we discretize each point $P_i$ (where subscript $i$ denotes the point index) onto a 3D integer grid with voxel size $\delta$, yielding integer coordinates $u_i = (u_{i,x}, u_{i,y}, u_{i,z})$. We define a bit-spreading operator $\mathcal{S}(v)$ that maps an integer $v$ to a sparse representation by shifting the $k$-th bit of $v$ to position $3k$. The Morton code $m_i$ is constructed by interleaving these spread bits as shown in the equation below:

\begin{equation}
    m_i = \mathcal{S}(u_{i,x}) \lor (\mathcal{S}(u_{i,y}) \ll 1) \lor (\mathcal{S}(u_{i,z}) \ll 2)
\end{equation}
where $\lor$ denotes the bitwise OR and $\ll$ denotes a left bit-shift. The final sequence is obtained by sorting the points such that $m_{\pi(1)} \le m_{\pi(2)} \le \dots \le m_{\pi(N)}$, where $\pi$ is the sorting permutation.

\subsection{Zero-Drift Quantized Compression}
\label{subsec:quantization}
% It is important to distinguish between prediction drift and quantization error. Traditional autoregressive models operating in continuous space often suffer from drift, where small floating-point prediction errors accumulate over the sequence, causing significant geometric distortion in later points.

% In contrast, 
LiZIP employs a \textit{Quantize-First} strategy. By snapping input points to a fixed integer grid before prediction, both the encoder and decoder operate on identical integer values. While this introduces a static, bounded quantization error (max $\sim$0.011 mm in our experiments), it guarantees \textbf{zero prediction drift}. Zero-drift refers to the elimination of cumulative prediction error across the auto-regressive sequence by calculating residuals in the integer domain; the 0.01mm error is a fixed, non-accumulating artifact of the initial quantization. The average reconstruction error of 0.0105 mm is approximately three orders of magnitude below the typical noise floor of automotive LiDAR sensors ($\pm$20 mm), rendering the compression functionally lossless for downstream perception tasks.

\subsection{Compressing Geometry (Neural Prediction)}
\begin{figure}[H]
    \centering
    \includegraphics[width=0.48\textwidth]{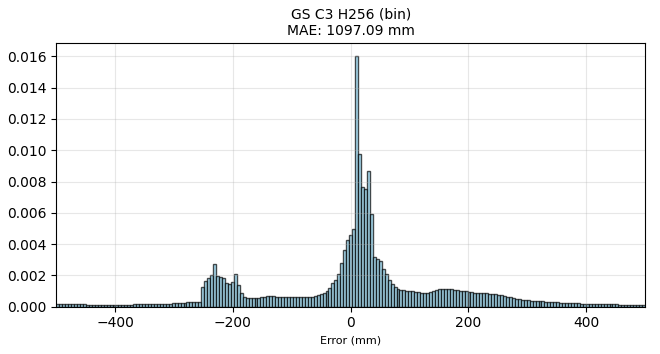}
    \caption{Distribution of prediction residuals by the chosen $k=3, H=256$ configuration after neural prediction. Residuals are sharply peaked around zero, which improves entropy coding efficiency.}
    \label{fig:residual_dist}
\end{figure}

Unlike LASzip which uses a second order linear predictor, LiZIP employs a Neural Predictive Model. For a target point $P_t$ (where subscript $t$ denotes the current point being compressed in the sequence), the compressor constructs a context vector $C_t$ from the $k = 3$ immediately preceding points. These points are normalized relative to $P_{t-1}$ to ensure translation invariance. The context is fed into the MLP with three hidden layers (256 neurons each, ReLU activation) \cite{Jiang2024TensorFlowMLP}. The MLP outputs a predicted coordinate triplet $\hat{P}_t$ where $\hat{P}_t$ is given by:

\begin{equation}
    \hat{P}_t = \text{MLP}(P_{t-3}, P_{t-2}, P_{t-1})
\end{equation}
The compressor then computes the residual $R_t = P_t - \hat{P}_t$. As shown in Fig.~\ref{fig:residual_dist}, the neural predictor yields residuals with a sharp Laplacian-like peak around zero, making them highly compressible by general-purpose entropy coders such as Zlib and LZMA.

\subsection{Residual Encoding and Byte Shuffling}
The floating-point residuals are quantized to 32-bit integers using a user-defined scaling factor (e.g., $10^5$ for micrometer precision). To compress these integers efficiently, LiZIP implements a \textit{Byte Shuffling} algorithm (also known as byte-plane transposition).

Formally, let $R = (r_1, \dots, r_N)$ be the sequence of $N$ quantized residuals. We represent each 32-bit integer $r_i$ as a tuple of four bytes $(b_{i,0}, b_{i,1}, b_{i,2}, b_{i,3})$, where $b_{i,0}$ denotes the least significant byte (LSB). The shuffling algorithm transposes this structure, creating a new stream $\mathcal{S}$ by concatenating the sequences of bytes at each significance position $k$:

\begin{equation}
    \mathcal{S} = \mathop{\Big\|}_{k=0}^{3} \left( b_{1,k}, b_{2,k}, \dots, b_{N,k} \right)
\end{equation}
where $\|$ denotes sequence concatenation. 

This transformation is critical for compression performance. In a standard array of 32-bit integers, data is stored as $[\text{Low}, \text{MidL}, \text{MidH}, \text{High}]$. Since the predictive residuals are typically small values clustered around zero, the high-order bytes ($b_{i,2}, b_{i,3}$) are predominantly zero. By grouping these bytes together, the transformed stream $\mathcal{S}$ contains long runs of identical zeros (e.g., \texttt{00 00 ...}), which dramatically lowers the entropy and allows the subsequent LZMA/Zlib coder to achieve higher compression ratios.

\subsection{Entropy Coding}
Finally, the shuffled byte stream is passed to the entropy coder. Unlike LASzip, which relies on a fixed custom arithmetic coder, LiZIP implements a modular backend allowing users to select between LZMA and Deflate (Zlib) depending on the application requirements.

\begin{itemize}
    \item \textbf{LZMA (Lempel-Ziv-Markov chain algorithm)}: This is the default configuration for maximum storage efficiency. By modeling high-order dependencies in the byte stream, LZMA achieves superior compression. 
    \item \textbf{Deflate (Zlib)}: For latency-critical applications, LiZIP can switch to the Deflate algorithm. While slightly less dense than LZMA, this configuration significantly reduces encoding/decoding time, making it suitable for real-time onboard processing where CPU cycles are a scarce resource.
\end{itemize}

\subsection{LiZIP File Format}
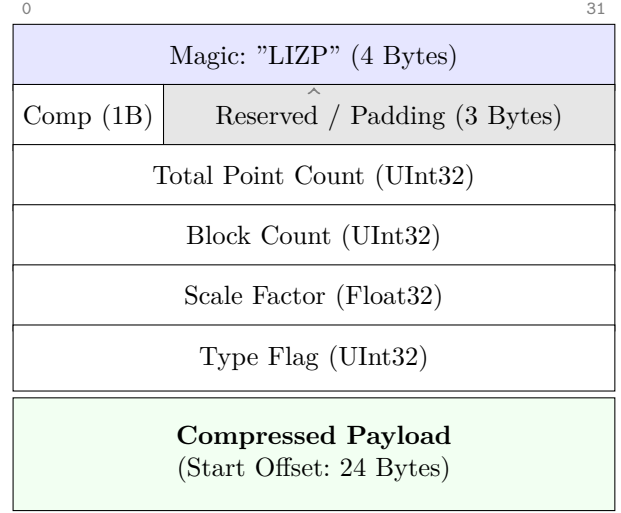
\begin{figure}[t]
\centering
\begin{tikzpicture}[
    byte/.style={rectangle, draw, fill=white, minimum height=2.5em, text centered},
    bitlabel/.style={font=\scriptsize\ttfamily, color=gray},
    header/.style={fill=gray!10, font=\bfseries}
]
    \def\w{8cm}
       \def\h{0.8cm}

       % --- HEADER (24 Bytes Total, 6 Rows of 4 Bytes) ---

       % Row 1: Magic (0-3 bytes)
       \node[byte, minimum width=\w, fill=blue!10] (magic) at (0,0) {Magic: "LIZP" (4 Bytes)};
       \node[bitlabel, above right] at (magic.north west) {0};
       \node[bitlabel, above left] at (magic.north east) {31};

       % Row 2: Comp ID (4) + Reserved (5, 6, 7)
       \node[byte, minimum width=0.25*\w, fill=white] (comp) at (-0.375*\w, -\h) {Comp (1B)};
       \node[byte, minimum width=0.75*\w, fill=gray!20] (res) at (0.125*\w, -\h) {Reserved / Padding (3 Bytes)};

       % Row 3: Total Point Count (8-11)
       \node[byte, minimum width=\w] (points) at (0, -2*\h) {Total Point Count (UInt32)};

       % Row 4: Block Count (12-15)
       \node[byte, minimum width=\w] (blocks) at (0, -3*\h) {Block Count (UInt32)};

       % Row 5: Scale Factor (16-19)
       \node[byte, minimum width=\w] (scale) at (0, -4*\h) {Scale Factor (Float32)};

       % Row 6: Type Flag (20-23)
       \node[byte, minimum width=\w] (type) at (0, -5*\h) {Type Flag (UInt32)};

       % --- PAYLOAD ---
       % Row 7: Payload (Starts at Byte 24)
       \node[byte, minimum width=\w, minimum height=1.5cm, fill=green!5, align=center] (payload) at (0, -6.6*\h) {
           \textbf{Compressed Payload} \\
           (Start Offset: 24 Bytes)
       };

       % Visual flow arrow
       \draw[->, thick, gray] (magic.south) -- (0, -0.5*\h);

\end{tikzpicture}
\caption{The LiZIP Binary File Format.}
\label{fig:file_structure}

% \vspace{-2.3em}
\end{figure}

To ensure storage efficiency and fast random access, LiZIP utilizes a custom binary file format. The file structure consists of a fixed length 24-byte header followed by a variable-length compressed payload. Figure \ref{fig:file_structure} illustrates the layout of the LiZIP file format.

\begin{itemize}
    \item \textbf{Magic Number (4 Bytes)}: The ASCII sequence $LIZP$ (Hex: \texttt{4C 49 5A 50}) identifies the file format.
    \item \textbf{Compression Flag (1 Byte)}: Indicates the entropy coding backend used (e.g., \texttt{0x02} for LZMA, \texttt{0x01} for Zlib).
    \item \textbf{Reserved (3 Bytes)}: Padding bytes used to align the subsequent 32-bit integers to memory boundaries.
    \item \textbf{Total Point Count (4 Bytes)}: An unsigned 32-bit integer specifying the total number of points.
    \item \textbf{Block Count (4 Bytes)}: A 32-bit unsigned integer specifying the number of compressed blocks.
    \item \textbf{Scale Factor (4 Bytes)}: A 32-bit float defining the quantization precision (e.g., $10^5$).
    \item \textbf{Type Flag (4 Bytes)}: A bitmask specifying the point cloud attributes included (e.g., $0x03$ indicates XYZ, Intensity).
    \item \textbf{Compressed Payload (Variable Length)}: Immediately following the header, the compressed payload begins. This section contains the LZMA or Zlib stream of the byte-shuffled residuals.
\end{itemize}

\section{Experimental Results}
We evaluated LiZIP on the NuScenes \cite{Caesar2020nuScenes} dataset. To ensure a consistent and deterministic baseline for model comparison, we utilized a standardized subset of the first 100 sequential LiDAR frames from each dataset. The evaluation frames consists of diverse urban environments across Boston, Singapore, Miami, and Pittsburgh, providing a representation of real-world LiDAR point clouds.

\subsection{Experimental Setup}
\begin{itemize}
    \item \textbf{Hardware:} All experiments were conducted on a workstation (Intel i7-13700H CPU, 16GB DDR5 RAM 5200MT/s) without any GPU acceleration to simulate the resource constraints of autonomous vehicle onboard systems.
    \item \textbf{Baselines:} We benchmarked against industry standards LASzip (specialized LiDAR compressor) and Google Draco (geometric KD-tree based compressor). For a general-purpose comparison, we also included GZip (Deflate) compression.
    \item \textbf{Metrics:} We measured compression ratio (original size / compressed size), File Size Reduction (\%), and Encoding/Decoding Latency (s).
\end{itemize}

\subsection{Training Methodology}
The neural predictor was trained on 80\% of 3,532 frames from the NuScenes dataset ($\sim$2,826 frames) and evaluated on 100 sequential frames randomly selected from the held-out 20\% test set. Training used a chunk-based approach (26 chunks of 140 frames, 80/20 train/validation split) with the Adam optimizer (an adaptive learning rate optimization algorithm that combines momentum and Root Mean Square (RMS) propagation) at a learning rate of $10^{-3}$ for 50 epochs per chunk. Training was performed on the same workstation described in the Experimental Setup (Intel i7-13700H with NVIDIA RTX 4060 GPU), requiring approximately 1 hour to complete.

\subsection{Model Selection and Architectural Efficiency}
All candidate models were first implemented and trained in Python (PyTorch) to leverage its rapid prototyping capabilities. To bridge the gap between training and high-performance inference, we developed a custom serialization protocol. A dedicated export script converts the PyTorch checkpoints (.pth) into a compact binary format (.bin). The exporter writes a 4-byte magic header (LIZM), followed by the network topology and the raw weights and biases stored as contiguous 32-bit floating-point arrays. This format strips away Python-specific metadata, allowing the C++ engine to mmap or load the neural parameters directly into memory with zero parsing overhead.

To determine the optimal predictive architecture, we performed a grid search over the Context Size ($k$) and Hidden Dimension ($H$) using our high-performance C++ implementation. This analysis was done with the goal of exploring the trade-offs between Context Size $k$ and Hidden Dimension $H$. We evaluated configurations across a grid of $k \in \{3, 5, 8\}$ and $H \in \{256, 512, 1024\}$

% \begin{table}[htbp]
%     \caption{Performance Comparison of LiZIP Neural Architectures (Python Implementation)}
%     \begin{center}
%     \begin{tabular}{|c|c|c|c|c|}
%     \hline
%     \multicolumn{2}{|c|}{\textbf{Architecture}}&\multicolumn{3}{c|}{\textbf{Performance Metrics}} \\
%     \cline{1-5}
%     \textbf{$k$} & \textbf{$H$} & \textbf{\textit{Latency (s)}}& \textbf{\textit{Size (KB)}}& \textbf{\textit{Max Error (mm)}} \\
%     \hline
%     3 & 256 & 0.44 & 201.70 & 0.011 \\
%     \hline
%     3 & 512 & 0.89 & 202.06 & 0.011 \\
%     \hline
%     3 & 1024 & 1.86 & 203.18 & 0.011 \\
%     \hline
%     5 & 256 & 0.50 & 206.65 & 0.011 \\
%     \hline
%     5 & 512 & 0.83 & 204.81 & 0.011 \\
%     \hline
%     5 & 1024 & 1.75 & 203.25 & 0.011 \\
%     \hline
%     8 & 256 & 0.47 & 206.11 & 0.011 \\
%     \hline
%     8 & 512 & 0.80 & 207.84 & 0.011 \\
%     \hline
%     8 & 1024 & 1.73 & 206.19 & 0.011 \\
%     \hline
%     \multicolumn{5}{l}{$k$: Context Size (Receptive Field), $H$: Hidden Layer Dimension.}
%     \end{tabular}
%     \label{tab:neural_configs}
%     \end{center}
% \end{table}

\begin{table}[htbp]
    \caption{Performance Comparison of LiZIP Neural Architectures (C++ Implementation, Average Total Latency per Frame)}
    
    \centering 
    
    \footnotesize
    \setlength{\tabcolsep}{2.5pt}

    \begin{tabular}{ccccc}
    \toprule
    \multicolumn{2}{c}{\textbf{Arch.}} & \multicolumn{3}{c}{\textbf{Metrics}} \\
    \cmidrule(r){1-2}
    \cmidrule(l){3-5}
    $k$ & $H$ & Lat. (s) & Size (KB) & Err. (mm) \\
    \midrule
    3 & 256 & 0.19 & 185.41 & 0.010 \\
    3 & 512 & 0.31 & 186.13 & 0.010 \\
    3 & 1024 & 1.06 & 185.42 & 0.010 \\
    5 & 256 & 0.18 & 186.17 & 0.010 \\
    5 & 512 & 0.33 & 184.88 & 0.010 \\
    5 & 1024 & 1.03 & 185.49 & 0.010 \\
    8 & 256 & 0.19 & 185.73 & 0.010 \\
    8 & 512 & 0.37 & 185.53 & 0.010 \\
    8 & 1024 & 1.23 & 184.50 & 0.010 \\
    \bottomrule
    \multicolumn{5}{l}{$k$: Context Size (Receptive Field), $H$: Hidden Layer Dimension.}
    \end{tabular}
    
    \label{tab:neural_configs_cpp_final}

    % \vspace{-1.5em} 

\end{table}

\begin{enumerate}
    \item \textbf{Impact of Hidden Dimension:}
    As shown in Table \ref{tab:neural_configs_cpp_final}, increasing the model capacity yields diminishing returns. While the largest model ($H=1024$) achieves the smallest absolute file size in some configurations (184.50 KB for $k=8$), the compression improvement relative to the baseline is negligible (less than 0.4\%). This minor gain comes at a significant computational cost. The inference latency increased by approximately \textbf{5.7$\times$} (0.18s vs 1.03s for $k=5$), rendering the larger models unsuitable for real-time operation.

    \item \textbf{Impact of Context Size:}
    Similarly, extending the context window beyond $k=3$ provided negligible benefits. All configurations cluster within a narrow range of 184.5-186.2 KB, demonstrating that spatial context beyond $k=3$ does not significantly improve compression. Given that the $k=3, H=256$ configuration achieves the lowest inference latency at \textbf{0.19s}, this represents the optimal trade-off between compression performance and computational efficiency.
\end{enumerate}

Based on this study, we identify the architecture with $k=3$ and $H=256$ as the optimal configuration. It achieves the lowest inference latency (0.19s) while maintaining compression performance within 1 KB of the absolute best-performing models. Consequently, all subsequent experimental results and comparisons against industry baselines are conducted using this specific configuration. The selected architecture ($k=3, H=256$) has a model footprint of 540 KB. The optimized C++ inference engine requires less than 50 MB of memory, ensuring compatibility with embedded automotive systems.

\subsection{Ablation Study: Component Efficiency}
\begin{figure}[t]
    \centering
    \includegraphics[width=\linewidth]{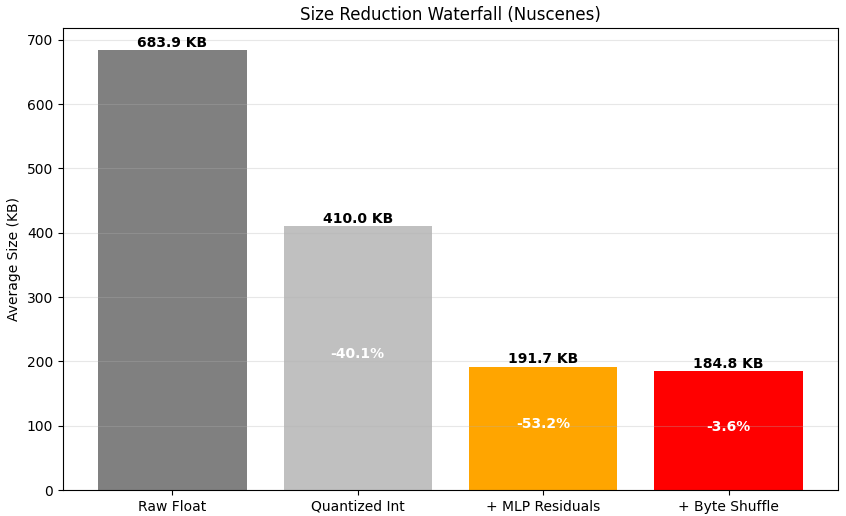}
    \caption{\textbf{Ablation Study of LiZIP Compression Stages.} The ``Waterfall'' chart highlights that the Neural Predictor (MLP) is the primary driver of compression (53.2\% reduction), distinguishing LiZIP from simple quantization schemes.}
    \label{fig:waterfall}
\end{figure}

To quantify the contribution of each stage in the LiZIP pipeline, we performed an ablation study on the NuScenes dataset. Figure \ref{fig:waterfall} illustrates the file size reduction at each processing step.

The transition from Raw Floats (683.9 KB) to Quantized Integers (410.0 KB) represents the standard removal of floating-point overhead. Crucially, the \textbf{Neural Prediction} stage serves as the primary compression engine, reducing the file size by a further 53.2\% (to 191.7 KB). This confirms that the network successfully learns the local geometric structure, rather than the gains being driven solely by quantization.

Finally, the \textbf{Byte Shuffling} stage contributes an additional 3.6\% reduction (to 184.8 KB) by reorganizing the residual bits to be more amenable to the LZMA entropy coder.

\section{Benchmarking Results}
We benchmarked the optimized C++ implementation ($k=3, H=256$, with LZMA and Zlib backends) against industry-standard compressors (GZip, LASzip, Draco). While the model was trained exclusively on the NuScenes dataset, we extended the evaluation to the unseen Argoverse dataset. The results demonstrate that LiZIP achieves higher compression ratios not only on the trained dataset but also outperforms all baselines on Argoverse, confirming the model's ability to generalize to diverse urban environments without retraining.

\subsection{Performance Benchmarks on NuScenes Dataset}
The results for the NuScenes dataset are summarized in Table \ref{tab:main_results}. LiZIP achieves a total file size reduction of approximately 71\% relative to the raw data and a further 7.5\% reduction beyond the industry-standard LASzip baseline. The original uncompressed size per frame is approximately 680 KB (based on NuScenes raw binary format of $\sim$34,000 points).

\begin{table}[htbp]
  \caption{Compression Performance on NuScenes (100 Frames)}
  \centering
%   \footnotesize
  \setlength{\tabcolsep}{5pt}
  \begin{tabular}{lccccc}
  \toprule
  Method & Enc & Dec & Size & vs. & Err \\
  & (ms) & (ms) & (KB) & LAZ & (mm) \\
  \midrule
  LiZIP (lzma) & 118$\pm$15 & 74$\pm$12 & 185.4$\pm$28 & -7.5\% & 0.010 \\
  LiZIP (zlib) & 42$\pm$8 & 33$\pm$5 & 198.2$\pm$29 & -1.1\% & 0.010 \\
  LASzip & 18$\pm$4 & 15$\pm$3 & 200.5$\pm$17 & --- & 0.011 \\
  Draco & 41$\pm$9 & 23$\pm$6 & 203.3$\pm$20 & +1.4\% & 0.033 \\
  GZip & 65$\pm$12 & 40$\pm$10 & 355.9$\pm$42 & +77.5\% & 0.000 \\
  \bottomrule
  \end{tabular}
  \label{tab:main_results}
\end{table}

While LiZIP incurs a higher computational cost compared to the heuristic-based LASzip, it offers a superior compression-density trade-off, making it ideal for bandwidth-constrained V2X streaming and long-term archival storage.

\begin{figure}[t]
    \centering
    \includegraphics[width=0.48\textwidth]{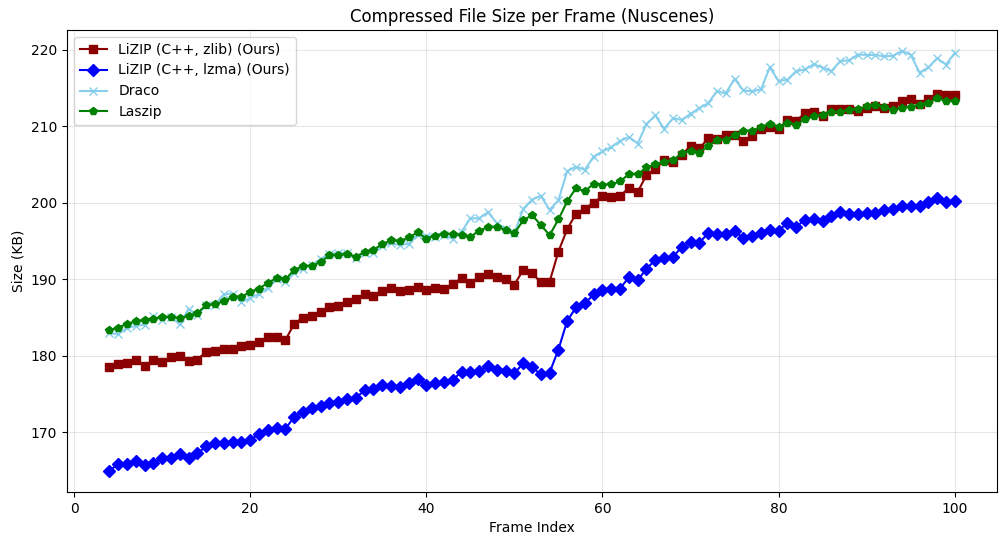}
    \caption{File Size Compression Breakdown of LiZIP against Baselines on NuScenes Dataset. Linear growth indicates consistent per-frame compression performance across all methods.}
    \label{fig:file_size_nuscenes}
\end{figure}

Figure~\ref{fig:file_size_nuscenes} shows the compression performance of LiZIP on the NuScenes dataset over 100 frames. The approximately linear growth across all baselines indicates stable, consistent per-frame compression with no error accumulation or degradation over the frames. LiZIP (LZMA) achieves 7.5\% reduction vs. LASzip while maintaining near-lossless quality (0.010 mm max error), demonstrating that neural predictive coding effectively captures urban LiDAR spatial correlations.

\subsection{Generalization to Argoverse Dataset}
To evaluate the model's generalization capabilities, we tested the NuScenes-trained model on the Argoverse dataset without any retraining or fine-tuning. The results are summarized in Table \ref{tab:argoverse_results}.

\begin{table}[htbp]
  \caption{Compression Performance on Argoverse (100 Frames)}
  \centering
  \setlength{\tabcolsep}{5pt}
  \begin{tabular}{lccccc}
  \toprule
  Method & Enc & Dec & Size & vs. & Err \\
  & (ms) & (ms) & (KB) & LAZ & (mm) \\
  \midrule
  LiZIP (lzma) & 255$\pm$25 & 160$\pm$20 & 602.3$\pm$6 & -14.8\% & 0.017 \\
  LiZIP (zlib) & 107$\pm$12 & 84$\pm$10 & 625.8$\pm$8 & -11.4\% & 0.017 \\
  LASzip & 28$\pm$5 & 23$\pm$4 & 706.5$\pm$6 & --- & 0.018 \\
  Draco & 56$\pm$8 & 31$\pm$5 & 679.2$\pm$6 & -3.9\% & 0.070 \\
  GZip & 38$\pm$6 & 24$\pm$4 & 973.5$\pm$13 & +37.8\% & 0.000 \\
  \bottomrule
  \end{tabular}
  \label{tab:argoverse_results}
\end{table}

\begin{figure}[t]
    \centering
    \includegraphics[width=0.48\textwidth]{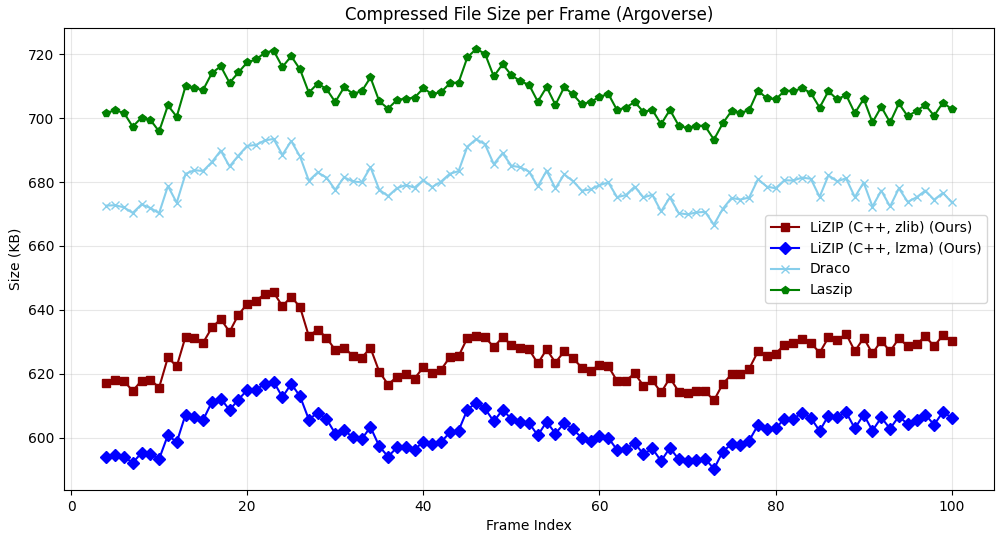}
    \caption{File Size Compression Breakdown of LiZIP against Baselines on Argoverse Dataset. Constant per-frame compression indicates uniform point cloud density across frames in this dataset.}
    \label{fig:file_size_argoverse}
\end{figure}

Figure~\ref{fig:file_size_argoverse} shows the compression performance of LiZIP on the Argoverse dataset over 100 frames. Notably, the Argoverse dataset exhibits more uniform point cloud sizes across frames compared to NuScenes, resulting in relatively constant per-frame compression ratios for all algorithms. This demonstrates that LiZIP's compression performance remains stable and predictable across datasets with different characteristics. The improved performance on Argoverse (14.8\% improvement vs. LASzip compared to 7.5\% on NuScenes) demonstrates that the learned geometric patterns transfer effectively across different LiDAR sensor configurations and urban environments. This suggests the neural predictor has captured fundamental 3D spatial correlations rather than dataset-specific artifacts.

\section{Discussion}
\label{sec:discussion}

The experimental results demonstrate that LiZIP successfully addresses the data bottleneck in autonomous vehicle sensor loops by balancing compression density with computational latency.

\subsection{Architectural Efficacy vs. Latency}
A crucial finding is that LiZIP (LZMA) consistently outperforms the industry-standard LASzip by 9.1\% on NuScenes \cite{Caesar2020nuScenes} and up to 14.8\% on the unseen Argoverse dataset \cite{Chang2019Argoverse}. This performance gap highlights the limitation of fixed-function linear predictors used in LASzip, while efficient for simple aerial scans, they fail to capture the complex, non-linear geometric structures of dense urban environments \cite{Isenburg2013LASzip}.

In terms of runtime, the LiZIP (zlib) configuration achieves an average decoding latency of 78ms. While this is higher than LASzip (33ms), it remains strictly within the 100ms real-time cycle required for 10Hz LiDAR sensors.

\subsection{Generalization}
Most learning-based compressors suffer from overfitting, performing poorly on datasets they were not trained on. However, LiZIP demonstrated superior performance on the Argoverse dataset (-14.8\% vs. LASzip) compared to the NuScenes training domain (-7.5\%). This counter-intuitive result suggests that the neural predictor has learned the geometric principles, such as the spatial correlation of road surfaces and buildings rather than memorizing dataset-specific sensor artifacts. This generalization is essential for V2X applications where vehicles may encounter diverse environments.

\subsection{Comparison with State-Of-The-Art Neural Frameworks}
While direct quantitative comparison is challenging due to differences in datasets (ScanNet/KITTI vs. NuScenes/Argoverse) and hardware configurations, it is critical to contextualize LiZIP with the broader landscape of neural point cloud compression. Table \ref{tab:neural_comparison} summarizes key metrics of recent neural baselines relative to LiZIP.

\begin{itemize}
    \item \textbf{Geometry Compression and Hardware Dependency:} Existing geometry compression frameworks such as VoxelContext-Net \cite{Que2021VoxelContextNet} and OctSqueeze \cite{Huang2021OctSqueeze} demonstrate impressive compression gains. VoxelContext-Net achieves a 43.66\% bitrate reduction over G-PCC on ScanNet \cite{Dai2017scannet}\cite{Que2021VoxelContextNet}. However, these results rely on high-end hardware acceleration (NVIDIA 2080Ti). In contrast, LiZIP operates entirely on a standard CPU (Intel i7). While OctSqueeze achieves faster decoding (6-8ms), it requires GPU parallelism to do so. In contrast, LiZIP operates entirely on a standard CPU. As shown in Table \ref{tab:main_results}, the LiZIP (zlib) configuration achieves a total pipeline latency (encoding + decoding) of approximately 75ms (42ms encode, 33ms decode). This allows the full compression-transmission-decompression cycle to complete within the 100ms real-time envelope of standard 10Hz LiDAR sensors, all without requiring expensive, power-hungry GPUs. This makes LiZIP uniquely suitable for the resource constraints of current embedded automotive controllers.
    \item \textbf{Lossless vs. Lossy Constraints:} Recent methods like RCPCC \cite{Cao2025RealtimeLiDARCompression} focus on lossy compression for robotics, achieving high compression rates (40x - 80x) and fast decoding (11.35ms) on CPUs \cite{Liu2021PointCloudAnalysis}. However, this comes at the cost of geometric precision, utilizing surface fitting that may smooth over critical small obstacles. LiZIP targets the "near-lossless" niche (max error $\sim$0.01mm), ensuring that the raw data integrity required for safety-critical long-range detection is preserved.
    \item \textbf{Attribute vs. Geometry:} MNeT \cite{Nguyen2023PointCloudCompression} outperforms G-PCC v11 by 8.4\% in attribute compression but focuses exclusively on color/intensity. LiZIP complements such approaches by focusing on the dominant data burden in LiDARs: spatial geometry $(x, y, z)$.
\end{itemize}

\begin{table}[h]
  \caption{Qualitative Comparison of Neural Compression Frameworks}
  \label{tab:neural_comparison}
  \centering
  \small
  \setlength{\tabcolsep}{2pt}
  \begin{tabular}{p{2cm}cccc}
  \toprule
  \textbf{Framework} & \textbf{Target} & \textbf{Hardware} & \textbf{Approach} & \textbf{Perf.} \\
  \midrule
  VoxelContext-Net & Geometry & GPU (2080Ti) & Octree+DL & -43.7\% \\
  OctSqueeze & Geometry & GPU (2080Ti) & Octree+Ent. & -15.0\% \\
  RCPCC & Geometry & CPU (i7) & Lossy Fit. & $\sim$60$\times$ \\
  MNeT & Attributes & GPU (3090) & Probabilistic & -8.4\% \\
  \midrule
  LiZIP (Ours) & Geometry & CPU (i7) & Neural Pred. & -14.8\% \\
  \bottomrule
  \end{tabular}
\end{table}

\subsection{Future Works}
While the current implementation demonstrates the viability of CPU-based neural compression, two key directions for future research remain.

First, we aim to port the inference engine to embedded GPU accelerators. The current MLP architecture is highly parallelizable, implementing the forward pass using NVIDIA TensorRT or CUDA kernels on edge platforms like the Jetson AGX Orin could potentially reduce inference latency from $\sim$40 ms to sub-millisecond levels, freeing up the CPU entirely for high-level planning tasks.

Second, we plan to integrate semantic awareness into the compression pipeline. By incorporating semantic segmentation labels (e.g., distinguishing 'pedestrians' from 'road surface'), the network could dynamically adjust its quantization precision. This would allow for aggressive compression of non-critical background elements while preserving sub-millimeter fidelity for safety-critical obstacles, further optimizing the bandwidth-accuracy trade-off for autonomous systems.

\section{Conclusion}
\label{sec:conclusion}
This paper presented LiZIP, a lightweight, near-lossless compression framework designed for the specific constraints of autonomous robotic systems. By combining Neural Predictive Coding with Morton-order spatial sorting, LiZIP successfully bridges the gap between deep learning capability and embedded system efficiency. Our evaluation on the NuScenes and Argoverse datasets demonstrates that LiZIP outperforms industry-standard algorithms, reducing file sizes by 7.5\%--14.8\% compared to LASzip and 8.8\%--11.3\% compared to Google Draco. Crucially, we achieved this performance on standard CPU hardware with total pipeline latencies ($\sim$75 ms) suitable for real-time operation. These results validate that learned geometric correlations can offer a superior alternative to fixed-function predictors for next-generation LiDAR data infrastructure.

% \section*{Acknowledgment}

% The preferred spelling of the word ``acknowledgment'' in America is without 
% an ``e'' after the ``g''. Avoid the stilted expression ``one of us (R. B. 
% G.) thanks $\ldots$''. Instead, try ``R. B. G. thanks$\ldots$''. Put sponsor 
% acknowledgments in the unnumbered footnote on the first page. 

\bibliographystyle{IEEEtran}
\bibliography{references}

\end{document}